\documentclass[11pt]{article}
\usepackage{geometry}
\usepackage{coling2020}
\usepackage{times}
\usepackage{url}
\usepackage{latexsym}
\usepackage{microtype}

\usepackage[hidelinks]{hyperref}

\colingfinalcopy 

\title{RGCL at SemEval-2020 Task 6: Neural Approaches to Definition Extraction}

 \author{Tharindu Ranasinghe$^1$, Alistair Plum$^1$, Constantin Or\u{a}san$^2$, Ruslan Mitkov$^1$ \\
 $^1$Research Group in Computational Linguistics, University of Wolverhampton, UK \\
 $^2$Centre for Translation Studies, University of Surrey, UK \\ 
  {\tt \{tharindu.ranasinghe, a.j.plum, r.mitkov\}@wlv.ac.uk} \\ 
  {\tt  c.orasan@surrey.ac.uk } \\}
  
\date{}

\begin{document}
\maketitle
\begin{abstract}
  This paper presents the RGCL team submission to SemEval 2020 Task 6: DeftEval, subtasks 1 and 2. The system classifies definitions at the sentence and token levels. It utilises state-of-the-art neural network architectures, which have some task-specific adaptations, including an automatically extended training set. Overall, the approach achieves acceptable evaluation scores, while maintaining flexibility in architecture selection.
\end{abstract}

\section{Introduction}

\blfootnote{
    \hspace{-0.65cm}  
    This work is licensed under a Creative Commons 
    Attribution 4.0 International Licence.
    Licence details:
    \url{http://creativecommons.org/licenses/by/4.0/}.
}

Definition Extraction refers to the task in Natural Language Processing (NLP) of detecting and extracting a \textit{term} and its \textit{definition} in different types of text. A common use of automatic definition extraction is to help building dictionaries \cite{Kobylinski2008}, but it can be employed for many other applications. For example, ontology building can benefit from methods that extract definitions \cite{Hearst1992,Malaise2007}, whilst the fields of definition extraction and information extraction can employ similar methodologies. It is therefore normal that there is growing interest in the task of definition extraction.

This paper describes our system that participated in two of the three subtasks of Task 6 at SemEval 2020 (DeftEval), a shared task focused on definition extraction from a specialised corpus. Our method employs state-of-the-art neural architectures in combination with automatic methods which extend and clean the provided dataset.


The remaining parts of this paper are structured as follows. First, we present related work in the area of definition extraction and the related field of relation extraction (Section \ref{sec:relwork}). The three subtasks and the dataset provided by the task organisers are described in Section \ref{sec:task}. Next, we describe our system (Section \ref{sec:meth}), followed by the results of the evaluation (Section \ref{sec:res}) and a final conclusion (Section \ref{sec:conc}).

\section{Related Work}
\label{sec:relwork}
The first efforts related to definition extraction happened in the field of hypernym extraction, where relations that usually indicate a definition were also dealt with. This includes the \textit{X is a type of Y} relation, such as  \textit{salmon is a type of fish}, where \textit{salmon} is a \textit{hyponym} of \textit{fish}.
Notable work includes \newcite{Hearst1992}, who automatically extracts hyponyms from large amounts of unstructured text using lexico-syntactic patterns. Inspired by this approach, \newcite{Malaise2004} describe a similar method to mine definitions in French, which are then classified in terms of their semantic relations, limited to the \textit{hypernymy - synonymy} relation. The approach is also used for building ontologies \cite{Malaise2007}.

The importance of the semantic relations between words for pattern-based approaches to definition extraction is highlighted in \cite{Sierra2008}. Here, the authors describe and explain definitional verbal patterns in Spanish, which they also propose to use for mining definitions. The proposed system is further presented in \newcite{Alarcon2009} and is aimed at Spanish technical texts. The system uses the aforementioned verbal patterns, as well as corresponding tense and distance restrictions, in order to extract a set of candidate terms and their definitions. Once extracted, the system applies some filtering rules and a decision tree to further analyse the candidates. Finally, the results are ranked using heuristic rules. All aspects of the system were developed by analysing the Institut Universitari de Ling\"{u}\'{i}stica Aplicada technical corpus in Spanish, which is also used for evaluation. 

Machine learning algorithms have also been used for definition extraction. \newcite{Gaudio2009} describe an approach that is said to be language independent and test it with decision trees and Random Forest, as well as Na\"{i}ve Bayes, k-Nearest Neighbour and Support Vector Machines using different sampling techniques to varying degrees of success. \newcite{Kobylinski2008} process Polish texts and use Balanced Random Forests, which bootstrap equal sets of positive and negative training examples to the classifier, as opposed to a larger group of unequal sets of training examples. Overall, while the approach is said to increase run time, it does bring minor increases in performance with some fine-tuning.

Most recently, \newcite{Spala2019} have created DEFT, a corpus for definition extraction from unstructured and semi-structured texts. Citing some of the pattern-based approaches also mentioned here, the authors argue that definitions have been well-defined and not necessarily representative of natural language. Therefore, a new corpus is presented that is said to more accurately represent natural language, and includes more messy examples of definitions. Parts of the DEFT corpus make up the dataset for this shared task, which is described in more detail in the following section.

\section{Subtasks and Dataset}
\label{sec:task}
The DeftEval shared task is split into three subtasks. The first is Sentence Classification, where the task is to predict whether a given sentence contains a definition. Subtask 2 is a sequence labelling task, which includes requires participants to assign BIO tags to indicate which tokens in a sentence belong to terms and their definitions. Furthermore, the BIO tags are fine-grained, denoting whether terms and  definitions are \textit{primary}, \textit{secondary} (the second time a term or definition has been seen in a text), \textit{referential} or \textit{ordered} (multiple terms that have inseparable definitions). The final subtask is Relation Classification which requires to classify the relation between terms and their definitions. Included are the tags \textit{direct} and \textit{indirect} definitions (links term or referential term to definition, respectively), \textit{supplements} (links indirect to direct definition), \textit{refers-to} (links referential term/definition to term/definition) and \textit{AKA} (links alias term to term).

The corpus provided by the organisers is made up of parts of the DEFT corpus described in \newcite{Spala2019}. This corpus has been compiled specifically for definition extraction tasks and is made up of legal contracts ($2.443$ sentences) and textbook data ($21.303$ sentences). Citing a growing need for definition extraction corpora, the creators also developed an annotation scheme that is specific to the task of definition extraction. 

\section{Methodology}
\label{sec:meth}
In this section we present the different approaches we employed for each subtask. The overall approach is based on a neural network architecture, but each subtask requires different methods of preprocessing the data, as well as task-specific tweaks to the data and architecture. Our implementation has been made available on Github.\footnote{\url{https://github.com/tharindudr/defteval}}

\subsection{Sentence Classification}
This section describes the methodology employed for \emph{Subtask 1: Sentence Classification}, as well as the experiments carried out in order to boost performance. We first present the methods used to process and extend the data, followed by a description of the main neural network architecture employed.

\subsubsection{Data Processing and Cleaning}
We first used the data converting python script that the organisers provided to convert the deft corpus in to classification instances. After that we concatenated all the files in the training folder in to single file and used it for training purposes while the concatenated file from the dev folder is used for evaluation purposes. 
As the Sentence classification task required only to predict 1 (contains a definition) or 0 (does not contain a definition) it was feasible to perform some simple cleaning to increase the classification performance without causing any side effects.
Upon analysis of the data we found that many sentences had some kind of numbering at the beginning, such as in the following example: 
\begin{quote}
    \textit{41. The evolution of various life forms on Earth can be summarized in a phylogenetic tree ([link])}
\end{quote}
Using a simple regular expression to match numbers and a punctuation mark at the beginning of a sentence, we removed these character strings across all sets. We used the same approach for finding and deleting character strings such as \textit{([link])}, which have been inserted by the task organisers to replace actual links to websites (see also the above example). In cases where the link replacement formed part of the sentence we did not perform a deletion:
\begin{quote}
    \textit{Examples of some neutral atoms and their electron configurations are shown in [link].}
\end{quote}
This decision was made as it would otherwise leave sentences incomplete.
After comparing the performance of our algorithm on both cleaned and uncleaned text we observed a marginal increase of $0.01$ across all evaluation metrics using on the best performing architecture. Other than this we did not carry out any additional cleaning. This was also due to the fact that we use BERT embeddings, making it unnecessary to remove any other characters, as it includes vectors for most characters.

\subsubsection{Data Augmentation}
In order to improve the performance of our classification we extend the training set automatically. To achieve this, the sequence labelling part of the system (described in Section \ref{sec:seqlab}) was used to detect terms in the training data. Where possible, we extracted the first sentence of the corresponding Wikipedia articles for these terms by scraping Wikipedia. This is due to the fact that the first sentence usually defines the term or item that the article is about. However, the approach had little impact on the performance of the system, trading increases in precision for decreases in recall and decreasing the F1-score by about 0.02. What we learned is since the data augmentation process is completely automated and not manually checked it introduces a certin level of noise to the dataset which result in decreasing the performance.  

\subsubsection{System Architecture}
In order to determine the most suitable system architecture for the sentence classification task, we experimented with three different neural architectures: Convolutional Neural Network (CNN) \cite{kim-2014-convolutional}, Recurrent Neural Network (RNN) \cite{10.1007/978-3-030-00018-9_15} and Transformer \cite{devlin2018bert}. After running various configurations, we found the Transformer architecture to perform best.

With the introduction of BERT \cite{devlin2018bert} transformer architectures have shown a massive success in a wide range of NLP tasks. Transformer architectures have been trained on general tasks like language modelling and then fine-tuned for classification tasks \cite{10.1007/978-3-030-32381-3_16,ranasinghe2019brums}. 

Transformer models take an input of a sequence and output the representation of the sequence. The sequence has one or two segments that the first token of the sequence is always [CLS] which contains the special classification embedding and another special token [SEP] is used for separating segments.

For text classification tasks, transformer models take the final hidden state \textbf{h} of the first token [CLS] as the representation of the whole sequence \cite{10.1007/978-3-030-32381-3_16}. The [CLS] token was then fed in to a simple softmax classifier to predict the label of the whole sentence: whether it contains a definition or not.



We fine-tuned all the parameters from transformer models as well as the softmax classifier jointly by maximising the log-probability of the correct label. For training the model, we used a batch-size of eight, Adam optimiser \cite{kingma2014adam} with learning rate $2\mathrm{e}{-5}$, and a linear learning rate warm-up over 10\% of the training data. The models were trained using only training data. Furthermore, they were evaluated while training using an evaluation set that had one fifth of the rows in training data. We performed early stopping if the evaluation loss did not improve over ten evaluation rounds. All the models were trained for three epochs. We experimented with several transformer architectures like BERT \cite{devlin2018bert}, XLNet \cite{yang2019xlnet}, XLM \cite{conneau2019unsupervised}, RoBERTa \cite{liu2019roberta} and DistilBERT \cite{sanh2019distilbert}. We used the HuggingFace's implementation of the transformer models \cite{Wolf2019HuggingFacesTS} and the pre-trained models available in the HuggingFace's model repository\footnote{\url{https://huggingface.co/models}}.

\subsection{Sequence Labelling}
\label{sec:seqlab}
This section describes the experiments we conducted for Subtask 2: \emph{ Sequence Labelling}. We first present the data processing methods used, followed by the neural network architecture employed. Due to the structure of the data and the way the annotations had to be made (CoNLL-like format) and evaluated, no cleaning was performed for this task.

\subsubsection{Data Processing and Augmentation}
As a preliminary step, we concatenated all the files from the train folder in Deft corpus to a single file and used it as the training file. Similarly we concatenated all the files from the dev folder in Deft corpus to a single file and used it for evaluation purposes.

For this subtask also we experimented with data augmentation techniques. We tried a similar approach as before, but with a bootstrapping focus: We used the classifier trained for this task to predict terms and extracted the first sentence from each corresponding Wikipedia article. Exploiting the structure of Wikipedia again, we simply automatically labelled the term in the corresponding sentence, therefore providing extra examples of the terms being used in a sentence. However, like in the previous case this step did not improve our results due to the noise it introduces. We also assume that the added terms were always mentioned at the beginning of a sentence, therefore adding positional bias to the classifier. 

\subsubsection{System Architecture}
We experimented with three different neural network architectures for the sequence labelling task: LSTM-CRF \cite{lample-etal-2016-neural}, Stack-LSTM  \cite{lample-etal-2016-neural} and Transformer  \cite{devlin2018bert}. In this task we also found that the Transformer architecture performs best. 


Transformer architectures have proved effective in NER tasks \cite{devlin2018bert}, which are also sequence labelling tasks. In light of this, in this subtask, we implemented the approach suggested in the first transformers paper - BERT \cite{devlin2018bert}: transformer model combined with a token-level classifier. After processing the sentence through the transformer model each word gets a vector representation. We used this vector representation as the input to the token-level classifier over the label set available for subtask 2. The token-level classifier consists of a dropout \cite{JMLR:v15:srivastava14a} and a linear classifier. We fine-tuned all the parameters from transformer models as well as the token-level classifier jointly by maximising the log-probability of the correct label.

For training the model, we used a batch-size of eight, Adam optimiser \cite{kingma2014adam} with learning rate $1\mathrm{e}{-5}$, and a linear learning rate warm-up over 6\% of the training data. The models were trained using only training data. Furthermore, they were evaluated while training using an evaluation set that had one fifth of the rows in training data. Similar to the subtask 1, we performed early stopping if the evaluation loss did not improve over ten evaluation rounds. All the models were trained for three epochs. We experimented with several transformer architectures: BERT \cite{devlin2018bert}, XLNet \cite{yang2019xlnet}, XLM \cite{conneau2019unsupervised}, RoBERTa \cite{liu2019roberta} and DistilBERT \cite{sanh2019distilbert}. We used the HuggingFace \textit{TokenClassification} interface \cite{Wolf2019HuggingFacesTS} and the pre-trained models available in the HuggingFace model repository\footnote{\url{https://huggingface.co/models}}.

We also experimented with adding a Conditional Random Field (CRF) layer \cite{10.1109/ICCV.2015.179} after the output of the Transformer. However evaluation of several configurations showed that adding the CRF layer does not improve the results. Therefore, we did not pursue these experiments any further. 



\section{Evaluation}
\label{sec:res}
In this section we present the evaluation results that were obtained during testing. We also provide a brief look at the final submission results of the shared task.

\subsection{Sentence Classification Results}
Table \ref{table:1} shows the evaluation of the different architectures we developed for the sentence classification task using the development set. We have also included baseline results which was performed using a Naive Bayes bag of words approach. It is clear that, while marginal, XLNet performs best overall. Interestingly, we compared BERT-Large against XLNet-Base, meaning that our best architecture was much less resource intensive to run.

For the final task evaluation using the test set, we achieved an F1-Macro score of $0.7885$, placing us 25th out of 56 participants. Compared to our evaluation results, this is a relatively high loss. We assume that our model has been largely overfitted in to the training set we used.

\begin{table*}[ht]
\centering
\begin{tabular}{l|ccc|ccc|ccc|c}
\hline
& \multicolumn{3}{c|}{\textbf{Not Definition}} & \multicolumn{3}{c|}{\textbf{Definition}} & \multicolumn{3}{c|}{\textbf{Weighted Average}} & \textbf{} \\ \hline
\multicolumn{1}{l|}{\textbf{Model}} & \textbf{P} & \textbf{R} & \textbf{F1} & \textbf{P} & \textbf{R} & \textbf{F1} & \textbf{P} & \textbf{R} & \textbf{F1} & \textbf{F1 Macro} \\ \hline
\textit{CNN} & 0.78 & 0.73 & 0.72 & 0.76 & 0.71 & 0.75 & 0.74 & 0.77 & 0.74 & 0.76 \\
\textit{RNN-BILSTM} & 0.76 & 0.71  & 0.74 & 0.68 & 0.74 & 0.72 & 0.75 & 0.72 & 0.73 & 0.75 \\
\textit{BERT} & 0.90 & 0.88 & 0.89 & 0.81 & 0.79 & 0.80 & 0.86 & 0.86 & 0.86 & 0.84 \\
\textit{XLNet} & 0.91 & 0.90 & 0.90 & 0.82 & 0.80 & 0.81 & 0.87 & 0.88 & 0.87 & \textbf{0.86} \\ \hline
\textit{Baseline} & 0.89 & 0.54 & 0.68 & 0.49 & 0.87 & 0.63 & 0.66 & 0.68 & 0.66 & 0.67 \\ \hline
\end{tabular}
\caption[Results for Subtask 1]{Results for Subtask 1 For each model, Precision (P), Recall (R), and F1 are reported on all classes, and weighted averages. Macro-F1 is also listed (best in bold).}
\label{table:1}
\end{table*}

\subsection{Sequence Labelling Results}
Table \ref{table:2} shows the evaluation results for the different architectures we tested for the Sequence Labelling task. As before, we see XLNet with the best results, and again see that the less resource intense base version is almost on par with the large version. It should also be noted that the best results were achieved with shortened maximum sequence lengths, down from 128 to 64.

In the official evaluation on the test set we ranked 28th of 51 with an F1-score of $0.4872$. This shows a significant drop in performance, possibly due to overfitting.

\begin{table}[ht]
\centering
\begin{tabular}{l|ccc}
\hline
\multicolumn{1}{l|}{\textbf{Model}} & \textbf{P} & \textbf{R} & \textbf{F1} \\ \hline
\textit{BERT} & 0.71 & 0.74 & 0.73 \\
\textit{ROBERTa} & 0.67 & 0.70  & 0.69 \\
\textit{XLNet - Base} & 0.71 & 0.75 & 0.73 \\
\textit{XLNet - Large} & \textbf{0.72} & \textbf{0.76} & \textbf{0.74} \\ \hline
\end{tabular}
\caption[Results for Subtask 2]{Results for Subtask 2 For each model, Precision (P), Recall (R), and F1 are reported overall (best in bold).}
\label{table:2}
\end{table}


\section{Conclusion}
\label{sec:conc}
We have presented the system the RGCL team has prepared for the SemEval-2020 Task 12. The design of the system allows for easy switching of different architectures to accommodate the needs of the task at hand. For this task, we have shown the Transformer architecture using XLNet is the most successful when working with limited resources. It has also been shown that data augmentation techniques we experimented, while not detrimental to overall performance, do not necessarily improve performance. In a shared task setting, the effect of the extended data from Wikipedia was not useful, however, for a wider approach with higher recall, this could be more helpful.

We also tried to participate in the final subtask, \emph{Relation Classification}. However, due to time constraints, we were not able to achieve a valid submission for the this subtask. We approached it as a sequence pair classification task and employed a Siamese Neural Network which was shown to perform well in sequence pair classification tasks \cite{10.5555/3016100.3016291,ranasinghe-etal-2019-semantic}. The architecture we employed is similar to the architecture presented in \cite{reimers-2019-sentence-bert}. When two sequences have a relation, we extracted the sequences and provided them as the input for the Siamese transformer architecture. Then we used the objective function suggested as classification objective function in \cite{reimers-2019-sentence-bert} and optimised the cross-entropy loss. Due to the complexity of this task, we managed to run only a baseline of the proposed architecture which achieved very low evaluation scores on the development data. Therefore, we did not have a submission for this task and do not present any results here. In future, we hope to carry out further experiments with Siamese transformer architectures for relation classification tasks.

Going forth, we also wish to use this system for further tasks across further languages. While we may not achieve the best performance, the system utilises realistic system resources and is therefore very versatile. This is particularly with regard to the first subtask, where the difference to the best team is around 0.09, whereas for subtask two the best team is 0.36 ahead of us, indicating that our system is not competitive. It is possible to extend these experiments to a different domain easily using a pretrained transformer model in that domain given that a corpus similar to deft corpus is available in that domain. For an example, our system should be easily adoptable to biology domain using the BioBERT pretrained transformer model \cite{10.1093/bioinformatics/btz682} and a deft corpus like corpus on biology domain. 

\bibliographystyle{coling}
\bibliography{semeval2020}

\end{document}